\documentclass[10pt,twocolumn,letterpaper]{article}

\usepackage[pagenumbers]{cvpr} 

\usepackage{graphicx, amsmath, amssymb, caption, subcaption, multirow, overpic, textpos}
\usepackage[table]{xcolor}
\usepackage[british, english, american]{babel}
\usepackage[pagebackref=false, breaklinks=true, letterpaper=true, colorlinks,
            citecolor=citecolor, linkcolor=linkcolor, bookmarks=false]{hyperref}
\definecolor{citecolor}{HTML}{0071BC}
\definecolor{linkcolor}{HTML}{ED1C24}

\def\cvprPaperID{**}
\def\confName{****}
\def\confYear{****}

\newlength\savewidth

\renewcommand{\paragraph}[1]{\vspace{1.25mm}\noindent\textbf{#1}}

\newcolumntype{x}[1]{>{\centering\arraybackslash}p{#1pt}}
\newcolumntype{y}[1]{>{\raggedright\arraybackslash}p{#1pt}}
\newcolumntype{z}[1]{>{\raggedleft\arraybackslash}p{#1pt}}

\newcommand{\app}{\raise.17ex\hbox{$\scriptstyle\sim$}}

\definecolor{deemph}{gray}{0.6}

\definecolor{baselinecolor}{gray}{.9}

\newcommand{\authorskip}{\hspace{2.5mm}}

\begin{document}
\title{
\vspace{-1mm}\Large GARNet: Global-Aware Multi-View 3D Reconstruction Network and \\the Cost-Performance Tradeoff\vspace{-3mm}}
\author{
 Zhenwei Zhu$^{1}$ \authorskip Liying Yang$^{1}$ \authorskip Xuxin Lin$^{1}$ \authorskip Chaohao Jiang $^{1}$ \authorskip Ning Li$^{1}$ \authorskip Lin Yang$^{2}$ \authorskip Yanyan Liang$^{1,}$\footnotemark \\[2mm]
 \small $^{1}$Macau University of Science and Technology, Faculty of Innovation Engineering \qquad $^{2}$Kingsoft, Seasun Games\\ \vspace{-4mm}
}
\maketitle

\renewcommand{\thefootnote}{*}

\footnotetext[1]{Corresponding author}

\begin{abstract}
Deep learning technology has made great progress in multi-view 3D reconstruction tasks. At present, most mainstream solutions establish the mapping between views and shape of an object by assembling the networks of 2D encoder and 3D decoder as the basic structure while they adopt different approaches to obtain aggregation of features from several views. Among them, the methods using attention-based fusion perform better and more stable than the others, however, they still have an obvious shortcoming — the strong independence of each view during predicting the weights for merging leads to a lack of adaption of the global state. In this paper, we propose a global-aware attention-based fusion approach that builds the correlation between each branch and the global to provide a comprehensive foundation for weights inference. In order to enhance the ability of the network, we introduce a novel loss function to supervise the shape overall and propose a dynamic two-stage training strategy that can effectively adapt to all reconstructors with attention-based fusion. Experiments on ShapeNet verify that our method outperforms existing SOTA methods while the amount of parameters is far less than the same type of algorithm, Pix2Vox++. Furthermore, we propose a view-reduction method based on maximizing diversity and discuss the cost-performance tradeoff of our model to achieve a better performance when facing heavy input amount and limited computational cost.
\end{abstract}

\section{Introduction}

3D reconstruction, a problem that involve the fields of computer vision and computer graphics, is considered as the core of many technologies such as computer-aided geometric design, computer animation, medical image processing, digital media and robotics. As a generation task, comparing to image restoration, lifting 2D images to 3D object is obviously an extremely difficult ill-posed inverse problem. According to the number of images as input, this task can be divided into single-view reconstruction \cite{wu2016learning,  fan2017point, wang2018pixel2mesh, tatarchenko2017octree, yang2021single} and multi-view reconstruction \cite{choy20163d, yang2020robust, xie2019pix2vox, wang2021multi}. In this work, we focus on the deep-learning-based algorithms which reconstruct the shape of 3D objects with voxel representation from multiple images.

\begin{figure}[!t]
	\centering
	\scalebox{1.22}{
	\includegraphics[width=0.8\linewidth]{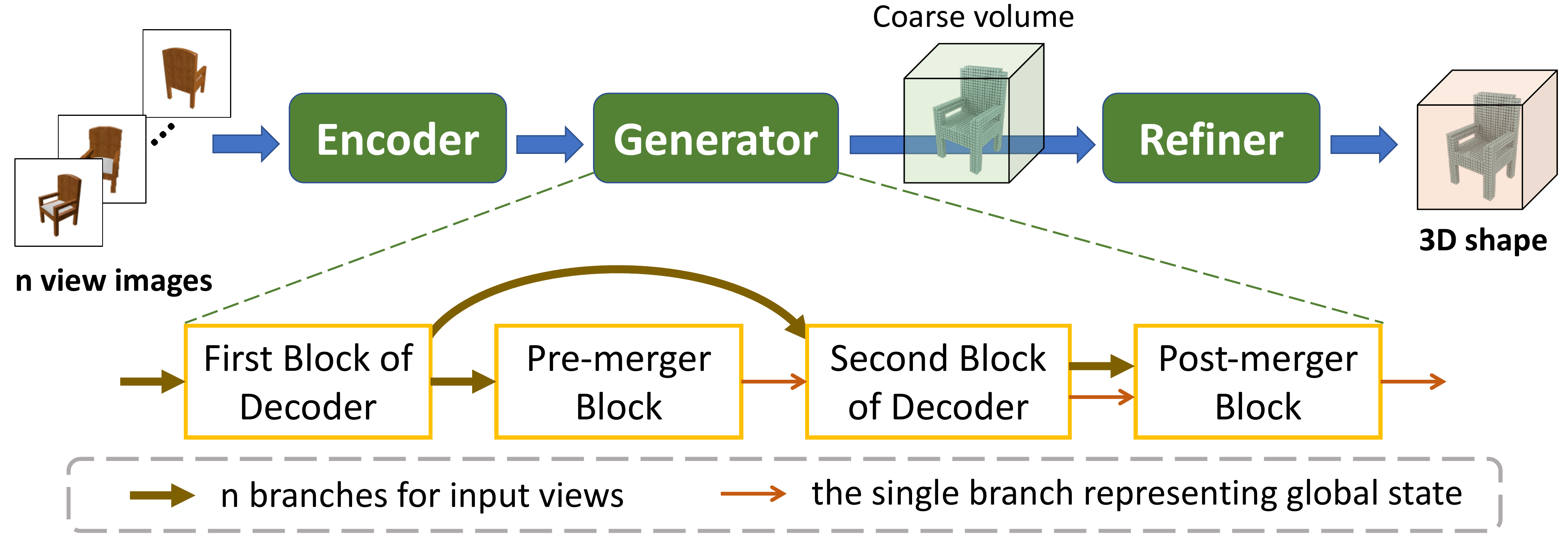}}
	\caption{Our network consists of encoder, generator and refiner. The data flow in generator reflects the concept of global-aware.}
	\label{highlight}
\end{figure}

At present, most mainstream solutions assemble 2D encoder and 3D decoder as a basic framework and reshape the high-level features between them as a two-dimensional connection to establish the mapping between an image and a voxel. Nevertheless, there is still an important issue for multi-view reconstruction — how to aggregate the features from an arbitrary number of views.

In our investigation, there are four types of fusion strategies identified. \cite{su2015multi, paschalidou2018raynet} adapt a pooling-based method to compress the feature map to a specific size using a pooling layer after concatenating the feature maps from all views. This dimensional collapse is too rough to avoid a massive loss of content. To make the fusion module learnable, 3D-R2N2 series \cite{choy20163d, ma2020improved} utilize recurrent neural network (RNN)-based methods. The features from all views are regarded as a sequence and processed by a recurrent unit before the decoder. However, it indicates inconsistent predictions for different permutations. In addition, such methods are not suitable for too many views as input because of the limited long-term memory. To address these shortcomings, attention-based fusion approaches create a sub-network to predict the confidence score map of each view and merge features based on it. Both AttSets \cite{yang2020robust} and Pix2Vox series \cite{xie2019pix2vox, xie2020pix2vox++} following this idea produce stable reconstructors. The former merges the features while the latter merges the voxels restored from each view directly. Very recent, some researches \cite{wang2021multi, yagubbayli2021legoformer} use transformer structure for multi-view reconstruction. Utilizing the natural advantage, the fusion process is integrated into the encoder stage. They perform well for a large number of view inputs, but the restoration quality is terrible when there are few input images.

We consider that the attention-based fusion performs better and more stable than the others, however, it still has an obvious shortcoming. During predicting the score maps, the connection of branches only relies on a softmax layer without any learnable parameters, so that it cannot be adaptive to the global state and only trusts the memory of the network. To address this issue, we propose a global-aware multi-view 3D reconstruction network, named GARNet, which not only follows the attention-based fusion but also establishes the correlation between each branch and the global. Figure~\ref{highlight} illustrates the composition of the network and highlights the data flow in the generator, which reflects the concept of global-aware fusion.

In practical problems, we provide a large number of view images for better results, however, the model may not process due to the constraints of time consumption and computational complexity. To overcome this trouble, we propose a view-reduction method based on maximizing diversity and discuss the cost-performance tradeoff of our model to achieve a better performance when facing heavy input amount and limited computational cost.

In detail, the contributions are as follows:
\begin{itemize}
\item\textbf{Network architecture:} We propose a global-aware fusion approach that inherits the stability of the attention-based mergers and establishes the correlation between views to achieve better performance.
\item\textbf{Loss function:} Precision and recall as quantitative indicators participate in the novel supervision to alleviate the weakness of expressing the difference between shapes brought by cross-entropy loss.
\item\textbf{Training strategy:} We propose a dynamic two-stage training strategy that distinguishes the network processes for single-view and multi-view input and alternates training in these two situations randomly. This strategy can effectively adapt to all reconstructors using attention-based fusion.
\item\textbf{Cost-performance tradeoff:} Facing heavy input amount, we provide a view-reduction method based on maximizing diversity to make the model achieve better performance under limited computational cost.
\end{itemize}

Furthermore, our models favorably against the SOTA methods \cite{xie2020pix2vox++, wang2021multi} in performance with fewer parameters than the same type of methods.

\section{Related Works}

\begin{itemize}
\item\textbf{Single-view 3D reconstruction.} In recent years, estimating 3D shape from a single view image is a hot topic. PointSetGeneration \cite{fan2017point} generates 3D shape based on the representation of point clouds. Pixel2Mesh \cite{wang2018pixel2mesh} represents the object by triangular mesh and process through a graph convolutional network (GCN). Voxel-based methods are rather widespread. \cite{girdhar2016learning} modifies voxel grids directly utilizing a 3DCNN. Using generative adversarial network (GAN) \cite{goodfellow2014generative}, 3DGAN \cite{wu2016learning} and 3DIWGAN \cite{smith2017improved} are proposed to solve the problem of 3D object generation, and the generator in these works can be converted to the single-view reconstructor by combining the variational auto-encoder (VAE) \cite{kingma2014auto}. For the high-resolution results, OGN \cite{tatarchenko2017octree} adapts the octree representation to overcome the trouble of a huge memory budget and designs a network to process it directly, however, Matryoshka Networks \cite{richter2018matryoshka} decompose a 3D shape into nested shape layers in a recursive manner. To bridge the gap between synthetic data and real-world data, DAREC \cite{pinheiro2019domain} and VPAN \cite{feng2021look} introduce the supervision on domain adaption during training. To supplement the missing information in the image, Mem3D \cite{yang2021single} constructs a memory network to offer the priors information accumulated from the training set.

\item\textbf{Multi-view 3D reconstruction.} SFM \cite{ozyecsil2017survey} and SLAM \cite{fuentes2015visual}, the traditional reconstruction approaches, rely on matching features to establish the relationship across different views, but they have great restrictions on usage scenarios. Recently, deep-learning-based approaches are popular in multi-view 3D reconstruction, frequently without viewpoint labels. \cite{lin2018learning} leverages 2DCNN to predict dense point clouds representing the surface of 3D objects. In Pixel2Mesh++ \cite{wen2019pixel2mesh++}, a coarse mesh can be improved iteratively with a series of deformations predicted by a GCN to form the final results. Representing by voxel, the methods focus on how to merge the features from several views. \cite{su2015multi, paschalidou2018raynet} compress the concatenating features from all views using a maximum pooling layer. 3D-R2N2 series \cite{choy20163d, ma2020improved} and LSM \cite{kar2017learning} receive the views one by one and extract the useful knowledge through a recurrent unit. EVolT\cite{wang2021multi} and LegoFormer \cite{yagubbayli2021legoformer} exploit the advantage of transformer structure to realize the blending of information between various views in the encoder stage. As the most stable methods currently, AttSets \cite{yang2020robust} and Pix2Vox series \cite{xie2019pix2vox, xie2020pix2vox++} apply the attention module on the multi-branch tasks, however, lack the exchange of information between branches.
\end{itemize}

\begin{figure*}[ht]
\centering
	\begin{minipage}[t]{0.9\linewidth}
	    \subfloat[]{
    		\includegraphics[width=1.0\linewidth]{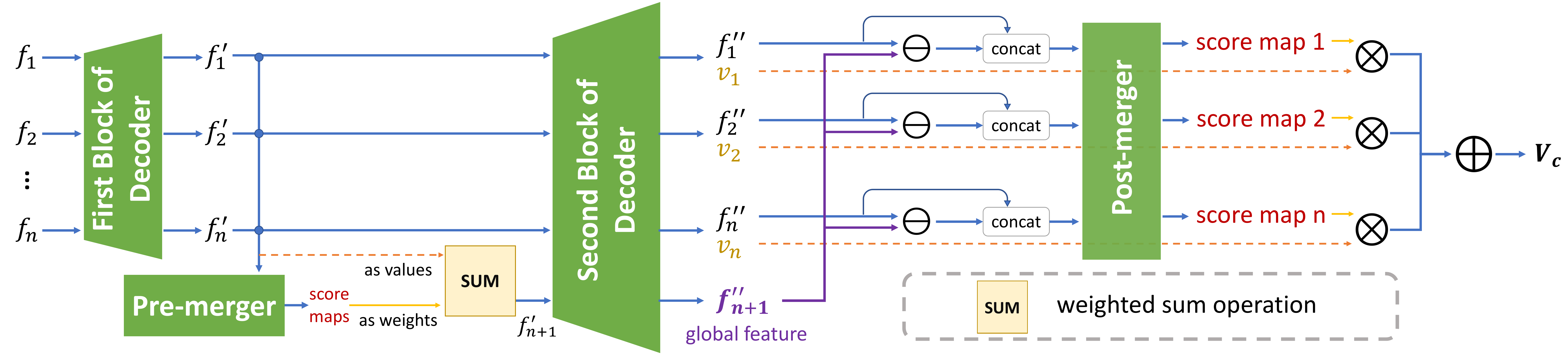}
    		\label{generator}}
    	\\
        \subfloat[]{
    		\includegraphics[width=1.0\linewidth]{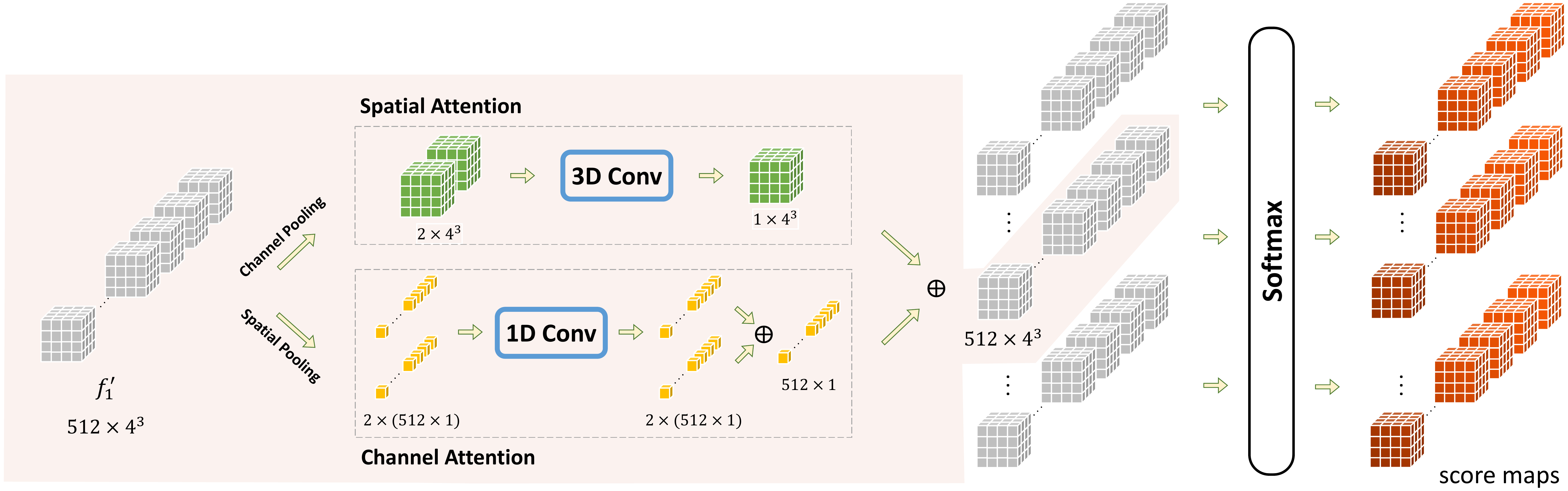}
    		\label{pre-merger}}
	\end{minipage}
	\caption{(a) is an overview of the generator that merges the features from different branches to reconstruct a coarse volume. It consists of two blocks of decoder, two fusion modules and some extra operations. Among them, the architecture of the pre-merger block, which predicts the score map for initial fusion, is shown as (b).}
\end{figure*}
\section{Methods}

According to this task, the goal of our architecture is to bridge an uncertain number of RGB images $ I = \{I_1, I_2, \cdots, I_n\}$ with a size of $ 224 \times 224 \times 3 $ to a binary voxel $V$ representing the shape with a size of $32\times32\times32$. Our network consists of three parts: encoder, generator and refiner. To begin, the encoder extracts feature map $f_i$ in parallel from each view image $I_i$. Then, the generator, which is formed by inserting two fusion blocks and a series of operations into a decoder, reconstructs a volume $V_c$ from these features exploiting the global-aware attention-based fusion that discusses in section~\ref{subsubsec::global-aware_fusion}. However, this volume still has great potentials for making further progress. Inspired by \cite{xie2020pix2vox++}, a refiner is adopted to modify $V_c$ to the final output $V$. Mathematically, the complete network is defined as:
\begin{equation}
	V = GARNet\left(I\right) = R\left(G\left(E\left(I_1, I_2, ..., I_n\right)\right)\right),
\end{equation}
where $E$, $G$ and $R$ denote the encoder, generator and refiner respectively. Referring to prior experience, encoder and decoder used in generator share the same structure as \cite{xie2020pix2vox++}. The refiner is a novel network, named 3D-U-ResNet, that combine the advantages of ResNet\cite{he2016deep} and U-Net\cite{ronneberger2015u}. It is a powerful structure with relatively lightweight and will be shown in the supplementary material.

\subsection{Global-Aware Fusion}
\label{subsubsec::global-aware_fusion}

In this section, we elaborate on the processing in generator that implements the global-aware fusion. As shown in Figure~\ref{generator}, the generator consists of four blocks, the two of which are split by the decoder and the others are pre-merger $M_{pre}$ and post-merger $M_{post}$. The decoder includes four transposed convolutional layers with a kernel size of $4^3$ and stride of $2$ and one transposed convolutional layer with a kernel size of $1^3$ and stride of $1$. The first layer is regarded as the first block of decoder, referred to as $D_1$, and the others compose the second block of decoder, referred to as $D_2$.

The input of the generator is feature maps extracted from view images. Considering to utilize the spatial information of features, we prefer to process the initial fusion on 4D tensors. However, these features derived from encoder lack the spatial relationship. Therefore, the first block of decoder happening before the initial fusion is reasonable.
\begin{equation}
	f_i^{\prime} = D_1 \left( E \left( I_i \right) \right) = D_1 \left( f_i \right) ,
\end{equation}
where $f_i^{\prime}$ is the output of $D_1$ on the branch of $I_i$.

Figure~\ref{pre-merger} shows the process of inferring the score map in pre-merger. Pre-merger employs two parallel attention mechanisms, a structure similar to the bottleneck attention module \cite{park2018bam}. Both maximum pooling and average pooling are utilized to compress features on channel and spatial to obtain a two-channel tensor and two vectors. The tensor is handled by a 3D convolution. For the high-dimensional vector, we replace the commonly fully connected layers with a 1D convolutional layer to maintain the structure lightweight inspired by \cite{2020ECA}. After the parallel module, the branch feature map for initial fusion is created by extending and adding the channel perception feature map and the spatial perception feature map. The features of all views are combined and processed by a softmax layer to predict the corresponding normalized score maps for each branch and each grid in $f^{\prime}$. Finally, the initial fusion feature map, referred to as $f_{n+1}^{\prime}$, is the result of the addition of $f^{\prime}$ weighted by the score maps and becomes the beginning of the $\left( n+1\right)th$ branch that need enter to $D_2$ like the other branches.
\begin{equation}
	f_{n+1}^{\prime} = \sum_i^n M_{pre} \left( f_i^{\prime} \right) \cdot f_i^{\prime}.
\end{equation}

The second block of decoder will reconstruct a voxel $v_i$ for each view. In addition, because the post-merger block needs to use the features of each branch to predict the score maps, we regard the concatenation of the feature maps from the last two layers of the decoder as the final features $f_i^{\prime\prime}$ of the view image $I_i$.
\begin{equation}
	\left( f_i^{\prime\prime}, v_i \right) = D_2 \left( f_i^{\prime} \right).
\end{equation}

According to the features of the $(n+1)$ branches, the post-merger block predicts the score maps of the first $n$ branches corresponding to the restored volumes from the $n$ view images. Since the actual meaning of the score map is the contribution rate of a branch to the entire, both the view and the global state are related to fusion logically. The input of post-merger for each branch is the concatenation of two feature maps: the feature map of the view $f_i^{\prime\prime}$ and the deviation between $f_i^{\prime\prime}$ and the global feature map $f_{n+1}^{\prime\prime}$ from the $\left( n+1 \right)$th branch. Due to only few channels of feature input, the post-merger, which consists of five 3D convolutional layers, has a more ordinary architecture than the pre-merger. Finally, following the same principle as pre-merger, the coarse volume is obtained after predicting the score maps with a softmax layer and fusing the reconstructed voxels from all view images weighted by the score maps.
\begin{equation}
	V_C = \sum_i^n M_{post} \left( f_i^{\prime\prime}, f_{n+1}^{\prime\prime} - f_i^{\prime\prime} \right) \cdot v_i .
\end{equation}

Summarizing the entire generator, we construct two attention-based fusion modules to implement global-aware. The first one merges all features to generate initial global features and then the second one makes full use of the global features to predict more reliable fusion weights for volumes.

\subsection{Loss Function}

The loss function is used to supervise both the coarse volume and the final volume. The binary cross-entropy (BCE) loss, which is commonly employed in previous works, is used to compare them to the ground truth respectively. Furthermore, we introduce precision and recall as new quantitative indicators to supervise the shape of the final volume. BCE only focuses on the classification of the cells but does not monitor overall shape. However, the precision and recall of occupied grids reflect the difference in shape. As a result, the combination can achieve the supervision to take both the local and the global into account. The complete loss function is defined as:
\begin{equation}
    L = \underbrace{\alpha L_{BCE\_V_{c}}}_{\text{for coarse volume}} + \underbrace{\beta L_{BCE\_V} + \gamma L_{Recall} + \mu L_{Precision}}_{\text{for fine volume}},
    \label{loss_function}
\end{equation}
where $\alpha$, $\beta$, $\gamma$, $\mu$ indicate the weights of each part. $L_{BCE\_V_{c}}$ and $L_{BCE\_V}$ represent the BCE loss function of the two reconstruction results and the following two items are the recall loss function and the precision loss function defined as:
\begin{equation}
    L_{Recall} = 1 - \frac{\sum_{i=1}^{32^3} p_{i} gt_{i}}{\sum_{i=1}^{32^3} gt_{i}},
\end{equation}
\begin{equation}
    L_{Precision} = 1 - \frac{\sum_{i=1}^{32^3} p_{i} gt_{i}}{\sum_{i=1}^{32^3} p_{i}},
\end{equation}
where $p$ and $gt$ denote the grids on the predicted result and ground truth.

\section{Dynamic Two-Stage Training Strategy}

At present, there are two strategies for training a reconstruction model using attention-based fusion. \cite{yang2020robust} proposes a feature-attention separate training (FASet) algorithm to spilt the training of the fusion module and the other parts into two stages. The first stage train all the parts except the fusion module using single-view input and the second stage only optimizes the fusion module using multi-view input. Pix2Vox series \cite{xie2019pix2vox, xie2020pix2vox++} offer another two-stage training strategy with a similar first stage to FASet while the second stage fed with random numbers of input to train the complete network, in which the random number is updated for each epoch. However, when the random number is 1, the fusion modules cannot get an useful backpropagation gradient limited by the softmax layer, which is proved in \cite{yang2020robust}. So, the second stage in Pix2Vox training is not reasonable enough. In addition, we consider that training the model sufficiently using single image input firstly just like the two mentioned strategies may easily lead to overfitting for the single-view reconstruction task and limit the performance for multi-view. 

To improve the performance of the attention-based fusion reconstructor, especially for multi-view input, we propose a general training strategy. First of all, the situations of single-view and multi-view as input are distinguished explicitly to perform different network processes. The propagation process does not go through the fusion module when the input is a single image, while for multi-view input, the parameters of the whole network participate. For each iteration, an integer less than or equal to the predefined value $n_{max}$ will be randomly chosen as the number of view inputs. As a result, the two modes, single-view reconstruction and multi-view reconstruction, alternate training at a certain vague frequency. It not only utilizes the stability brought by the former to assist the train of the latter but also avoids model overfitting trending to the former task. The method to integrate the two training modes in this way is named dynamic two-stage training strategy and its effectiveness is verified in Section~\ref{sec:training_strategy}.

\section{Cost-Performance Tradeoff via View-Reduction}

We have established a robust and stable multi-view 3D reconstruction network so far. In practice, users can employ more view images as input to achieve better reconstruction results. However, increasing input images means greater computational complexity and more time consumption. In some cases, the acceptable time for inference is strictly limited. It is necessary to provide a cost-performance tradeoff analysis for generating relatively high-quality results with limited computational cost. Considering the multi-view reconstruction accuracy relies on the diverse viewpoints information, we propose a view-reduction approach based on maximizing diversity to remove some branches and control the computational cost while preserving performance. Intuitively, a combination of views with a considerable disparity in viewpoints taking place of similar views can provide more diverse information for restoring a reliable volume. As a result, we attempt to find a combination of images with roughly complementary information from all view input to reduce the branches. The approach is based on isometric mapping training and maximizing diversity selection.

\begin{figure}[ht]
	\centering
	\scalebox{1.22}{
	\includegraphics[width=0.8\linewidth]{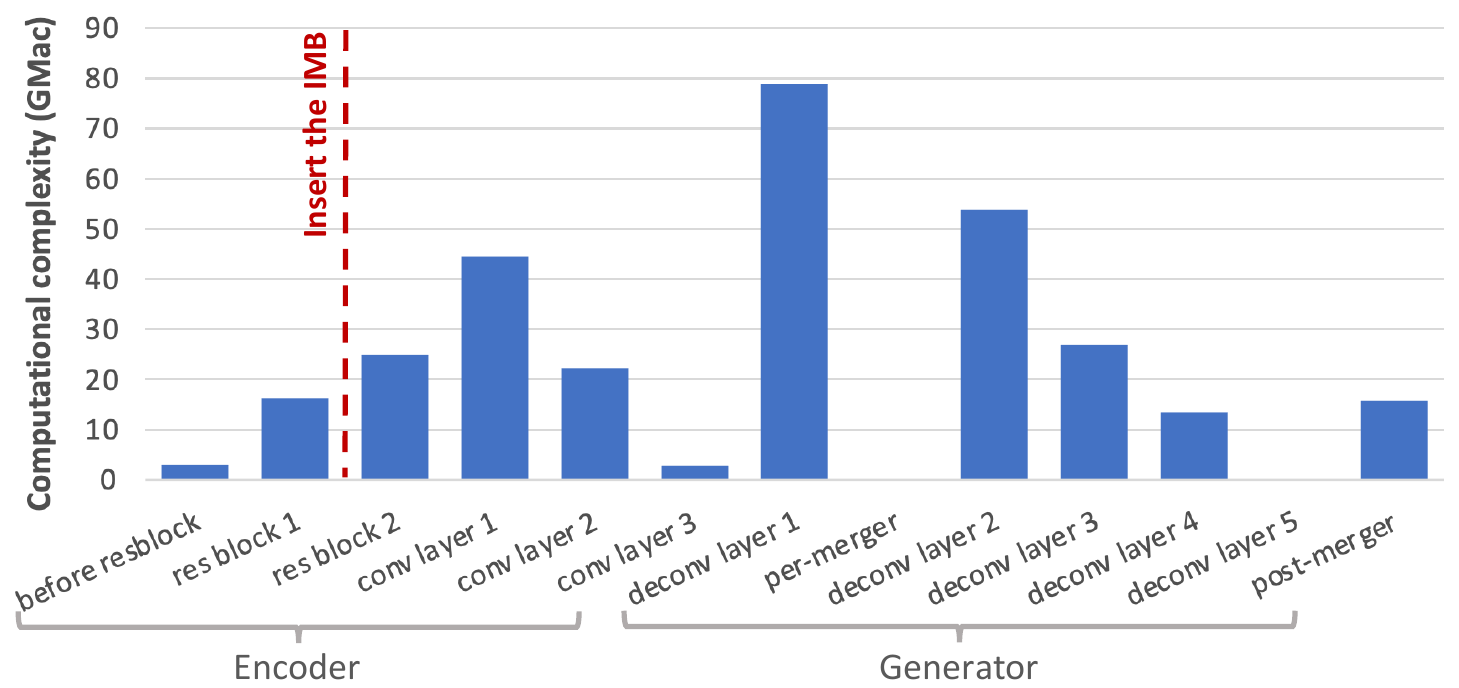}}
	\caption{The distribution of computational complexity in encoder and generator.}
	\label{computational_complexity_each_layer}
\end{figure}

\subsection{Isometric Mapping}

To extract the diverse viewpoint information via reducing the redundant features, we assume a low-dimensional manifold space in which each point represents observation position information matching to a view image of the object, and the Euclidean distance between two points is consistent with the difference of the viewpoints. Derived from the space, it is convenient to pick a subset of points with a widely scattered distribution that corresponds to an image combination with relatively complementary information.

\begin{table*}[]
	\centering
	\renewcommand\arraystretch{1.1}
	\scalebox{0.75}{
	\begin{tabular}{c|ccccccccc}
		\multicolumn{1}{l|}{} & \multicolumn{1}{c}{\textbf{1 view}} & \multicolumn{1}{c}{\textbf{2 views}} & \multicolumn{1}{c}{\textbf{3 views}} & \multicolumn{1}{c}{\textbf{4 views}} & \multicolumn{1}{c}{\textbf{5 views}} & \multicolumn{1}{c}{\textbf{8 views}} & \multicolumn{1}{c}{\textbf{12 views}} & \multicolumn{1}{c}{\textbf{16 views}} & \multicolumn{1}{c}{\textbf{20 views}} \\ \hline
		\textbf{3D-R2N2}\cite{choy20163d} & 0.560 / 0.351 & 0.603 / 0.368 & 0.617 / 0.372 & 0.625 / 0.378 & 0.634 / 0.382 & 0.635 / 0.383 & 0.636 / 0.382 & 0.636 / 0.382 & 0.636 / 0.383\\
		\textbf{AttSets}\cite{yang2020robust} & 0.642 / 0.395 & 0.662 / 0.418 & 0.670 / 0.426 & 0.675 / 0.430 & 0.677 / 0.432 & 0.685 / 0.444 & 0.688 / 0.445 & 0.692 / 0.447 & 0.693 / 0.448 \\
		\textbf{Pix2Vox++}\cite{xie2020pix2vox++} & 0.670 / \textbf{0.436} & 0.695 / 0.452 & 0.704 / 0.455 & 0.708 / 0.457 & 0.711 / 0.458 & 0.715 / 0.459 & 0.717 / 0.460 & 0.718 / 0.461 & 0.719 / 0.462 \\
		\textbf{EVolT}\cite{wang2021multi} & - / - & - / - & - / - & 0.609 / 0.358 & - / - & 0.698 / 0.448 & 0.720 / 0.475 & 0.729 / 0.486 & 0.735 / 0.492 \\
		\textbf{Legoformer}\cite{yagubbayli2021legoformer} & 0.519 / 0.282 & 0.644 / 0.392 & 0.679 / 0.428 & 0.694 / 0.444 & 0.703 / 0.453 & 0.713 / 0.464 & 0.717 / 0.470 & 0.719 / 0.472 & 0.721 / 0.472 \\ \hline
		\textbf{GARNet} & \textbf{0.673} / 0.418 & \textbf{0.705} / \textbf{0.455} & \textbf{0.716} / \textbf{0.468} & \textbf{0.722} / \textbf{0.475} & \textbf{0.726} / 0.479 & 0.731 / 0.486 & 0.734 / 0.489 & 0.736 / 0.491 & 0.737 / 0.492 \\
		\textbf{GARNet+} & 0.655 / 0.399 & 0.696 / 0.446 & 0.712 / 0.465 & 0.719 / \textbf{0.475} & 0.725 / \textbf{0.481} & \textbf{0.733} / \textbf{0.491} & \textbf{0.737} / \textbf{0.498} & \textbf{0.740} / \textbf{0.501} & \textbf{0.742} / \textbf{0.504} \\ \hline    
	\end{tabular}}
	\caption{Evaluation and comparison of the performance on ShapeNet using IoU / F-Score$@1\%$. The best results are highlighted in bold.}
\label{total_results}
\end{table*}

Therefore, we build an isometric mapping block (IMB) with a simple structure composed of two paralleled pooling layers and three MLPs (multilayer perceptron) for dimensionality reduction of encoded features, and then insert it into the reconstruction model to obtain a more compact representation preserving information of viewpoints. Such that we can use fewer view branches to reconstruct the volume based on maximizing diversity of viewpoints information. Figure~\ref{computational_complexity_each_layer} records the distribution of computational complexity before the refiner when facing a heavy input amount (24 views). We expect to assign fewer operations before IMB since the quantity of calculation can only be reduced after that by eliminating branches. As a result, it is reasonable to use the output of the first residual block in the encoder as the input of IMB. Training IMB still adopts the data without location and direction information of views. As mentioned by \cite{xie2019pix2vox}, the score maps for volumes predicted by the fusion module can be thought of as a representation of the visible parts from a viewpoint learned by the network adaptively. The weights allocated to the visible parts will presumably be higher. Consequently, the difference between the score maps predicted by the post-merger block in our model can be used to distinguish the difference between the visible parts, i.e. the relationship of viewpoints.

To train IMB, a manifold learning algorithm similar to \cite{tenenbaum2000global} is established for isometric mapping the low-dimensional points in Euclidean space to the score maps for different views of one object in L1 space. The L1 loss utilized to supervise is defined as:
\begin{equation}
    L_{IMB} = \left\| \sum_i^n \sum_j^n \left( \left\| P_i - P_j \right\|_{L^2} - \left\| Q_i - Q_j \right\|_{L^1} \right) \right\|_{L^{1}},
\end{equation}
where $P$ represents the points in low-dimensional manifold space corresponding to view images of an object and $Q$ denotes their matching score maps.

\subsection{Diversity Maximization}

The problem is now formulated as reducing $N$ branches of views to $n$. We obtain $N$ low-dimensional points via IMB. Using farthest point sampling (FPS), $n$ points can be selected and only their corresponding branches are retained to continue to finish the rest of the reconstruction network.

FPS is a sampling method with uniform and wide coverage so that the selected points maximize the diversity of the entire point set in a limited capacity. These points correspond to a view combination with a great variation in viewpoints. Meanwhile, the score maps, which indicate the contribution of views to different voxel grids, are directly tied to the meaning of these points. Either retaining branches with widely varying viewpoints or larger differences of contribution weights distribution imply maximizing diversity.

Thus, the view-reduction method for cost-performance tradeoff by retaining the view combination with more diverse information for reconstruction is realized and the  effectiveness will be verified in Section~\ref{sec:tradeoff}.

\begin{table}[]
	\centering
	\renewcommand\arraystretch{1.05}
	\scalebox{0.8}{
	\begin{tabular}{c|ccccc}
        \hline
		\textbf{} & \multicolumn{1}{c}{\textbf{1 view}} & \multicolumn{1}{c}{\textbf{2 views}} & \multicolumn{1}{c}{\textbf{3 views}} & \multicolumn{1}{c}{\textbf{4 views}} & \multicolumn{1}{c}{\textbf{5 views}} \\ \hline
		\textbf{Setup 1} & 0.670 & 0.695 & 0.704 & 0.708 & 0.711 \\
		\textbf{Setup 2} & 0.6693 & 0.6990 & 0.7090 & 0.7138 & 0.7172  \\
		\textbf{Setup 3} & 0.6707 & 0.7026 & 0.7136 & 0.7187 & 0.7223 \\ 
		\textbf{Setup 4} & 0.6697 & 0.7023 & 0.7137 & 0.7195 & 0.7235 \\
		\textbf{Setup 5} & \textbf{0.6725} & \textbf{0.7047} & \textbf{0.7160} & \textbf{0.7217} & \textbf{0.7255} \\
		\textbf{Setup 6} & 0.6551 & 0.6958 & 0.7117 & 0.7193 & 0.7246 \\\hline\hline
		\textbf{} & \multicolumn{1}{c}{\textbf{8 views}} & \multicolumn{1}{c}{\textbf{12 views}} & \multicolumn{1}{c}{\textbf{16 views}} & \multicolumn{1}{c}{\textbf{20 views}} & \\ \hline
		\textbf{Setup 1} & 0.715 & 0.717 & 0.718 & 0.719 & \\
		\textbf{Setup 2} & 0.7217 & 0.7245 & 0.7262 & 0.7269 & \\
		\textbf{Setup 3} & 0.7280 & 0.7308 & 0.7324 & 0.7332 & \\ 
		\textbf{Setup 4} & 0.7289 & 0.7319 & 0.7336 & 0.7345 & \\
		\textbf{Setup 5} & 0.7312 & 0.7340 & 0.7357 & 0.7368 & \\
		\textbf{Setup 6} & \textbf{0.7331} & \textbf{0.7373} & \textbf{0.7400} & \textbf{0.7415} & \\\hline
	\end{tabular}}
	\caption{The ablation experiments on ShapeNet about dynamic two-stage training strategy, 3D-U-ResNet as the refiner network, global-aware fusion, precision-recall loss function and 8-view input.}
\label{ablation_experiments}
\end{table}

\section{Experiments}

\begin{itemize}
\item\textbf{Dataset.} We evaluate our reconstruction network on the ShapeNet\cite{wu20153d} dataset using both Intersection of Union (IoU) and F-Score$@1\%$ \cite{tatarchenko2019single,xie2020pix2vox++} as the metric. Following \cite{choy20163d}, only a subset of ShapeNet including 13 categories and 43,783 3D objects with 24 randomly view images for each are used in our experiments.

\item\textbf{Implementation Details.} We adopt an Adam optimizer \cite{kingma2014adam} with $\beta_{1}=0.9$ and $\beta_{2}=0.999$ to train our multi-view reconstruction network with a batch size of 32 for 200 epochs, with 140 epochs using only BCE loss function and then 60 epochs using the complete loss function as mentioned previously. The weights in Equation~\ref{loss_function} are set as $\alpha=\beta=10$ and $\gamma=\theta=0.5$. The learning rate is 1e-3 initially and reduce to half after $\left[40, 60, 80, 100, 140, 180\right]$ epochs sequentially. For the property of dynamic two-stage training strategy, the parameters in the fusion module are optimized less frequently than the others. Thus, setting a slightly higher learning rate for the fusion module part can archive a better performance. Eventually, we provide two models respectively setting the maximum number of input views to 3 and 8 during training, named GARNet and GARNet+. The fixed threshold for binarizing the probabilities is set as 0.3. It takes about 2 days to train GARNet on 1 Tesla V100 and about 3 days to train GARNet+ on 2 Tesla V100. The IMB for view-reduction is extra trained for 15 epochs relied on a converged and frozen reconstruction network, and uses 40 objects with 24 views for each iteration. The learning rate is set to 2e-3 and decreased by 0.1 every 5 epochs.
\end{itemize}

\subsection{Multi-View Reconstruction Results}

\begin{figure*}[ht]
    \centering
	\begin{minipage}{0.81\linewidth}
        \centerline{\includegraphics[width=1\linewidth]{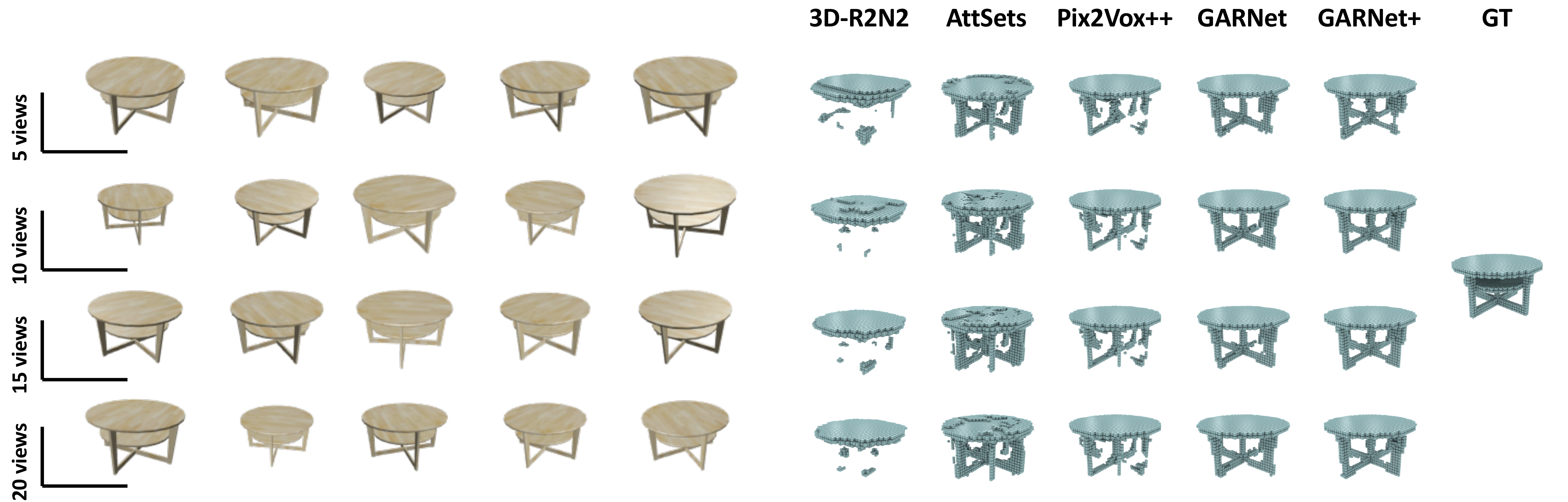}}
	\end{minipage}
	\\
	\begin{minipage}{0.81\linewidth}
        \centerline{\includegraphics[width=1\linewidth]{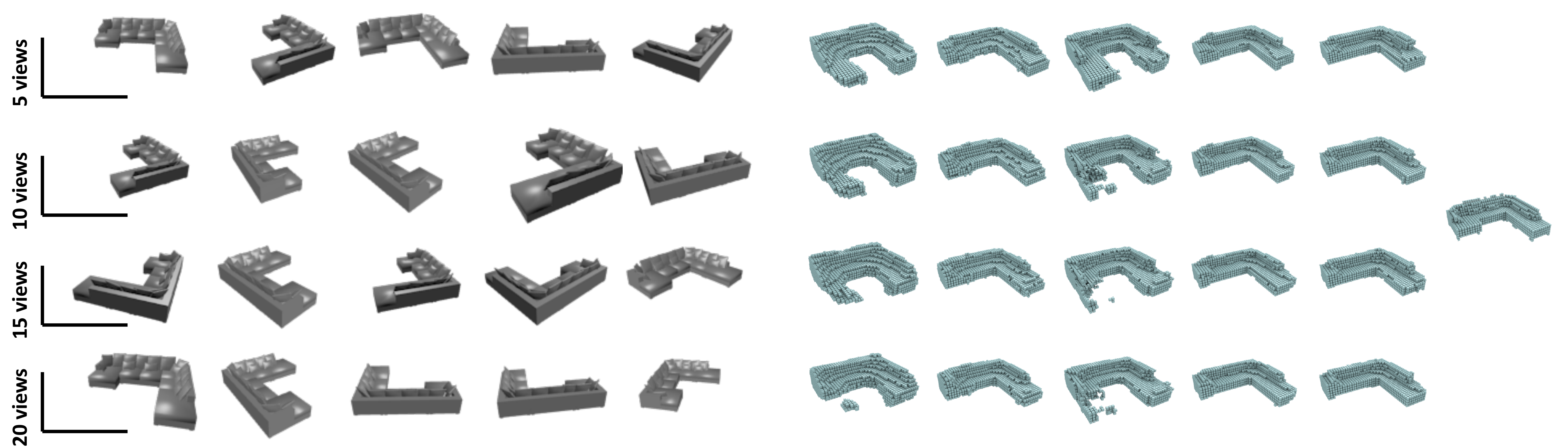}}
	\end{minipage}
	\caption{Multi-view reconstruction results on the test set of ShapeNet when facing 5 views, 10 views and 15 views as input.}
\label{show_results}
\end{figure*}

The performance of our models for multi-view reconstruction is evaluated and compared with existing SOTA methods. As shown in Table~\ref{total_results}, our proposed GARNet already dominates in almost all metrics. Not only does it performs consistently well when few views as input just like the other reconstructors using attention-based fusion, but also it breaks the advantage of the transformer method EvolT \cite{wang2021multi} when a large number of views as input. Furthermore, we also provide GARNet+, which performs better in multi-view reconstruction tasks while sacrificing acceptable performance for single-view. The two models are identical in structure and both of them outperform existing SOTA methods while only using about 69\% of parameters of the same type of algorithm, Pix2Vox++ \cite{xie2020pix2vox++}.

As examples, Figure~\ref{show_results} shows several reconstruction results. Comparing to the other methods, for table restoration, our models present a smoother plane and depict the links between the legs more accurately. In addition, our results for the sofa are also stable and reasonable, however, Pix2Vox++ produces a part of incorrect and confusing voxel girds. Without the global-aware, mistakes in a branch will seriously affect the final result. We exploit the fusion at the feature level, which is relatively insensitive to local perception, as the global information to bring the network a certain ability for self-correction.

\begin{table}[]
	\centering
	\renewcommand\arraystretch{0.95}
	\scalebox{0.8}{
	\begin{tabular}{c|rrrrrrrrr}
		\textbf{CA layer} & 
		\multicolumn{1}{c}{\textbf{1 view}} & \multicolumn{1}{c}{\textbf{2 views}} & \multicolumn{1}{c}{\textbf{3 views}} & \multicolumn{1}{c}{\textbf{4 views}} & \multicolumn{1}{c}{\textbf{5 views}} \\ \hline
		\textbf{FC} & 0.6652 & 0.6991 & 0.7111 & 0.7167 & 0.7203 \\
		\textbf{1D Conv} & \textbf{0.6697} & \textbf{0.7023} & \textbf{0.7137} & \textbf{0.7195} & \textbf{0.7235} \\ \hline\hline
		\textbf{CA layer} & 
		 \multicolumn{1}{c}{\textbf{8 views}} & \multicolumn{1}{c}{\textbf{12 views}} & \multicolumn{1}{c}{\textbf{16 views}} & \multicolumn{1}{c}{\textbf{20 views}} & \\ \hline
		 \textbf{FC} & 0.7259 & 0.7295 & 0.7313 & 0.7322 & \\
		\textbf{1D Conv} & \textbf{0.7289} & \textbf{0.7319} & \textbf{0.7336} & \textbf{0.7345} & \\ \hline
	\end{tabular}}
	\caption{Comparison of performance evaluated by IoU when using fully connection (FC) layers and a 1D convolutional layer in channel attention (CA) module of pre-merger. Experiments on ShapeNet using only BCE loss function and setting the upper limit of input views to 3 during training.}
\label{bam+eca}
\end{table}

\subsection{Ablation Experiments}

We use Pix2Vox++ \cite{xie2020pix2vox++} as the baseline. To design ablation experiments, our proposed approaches, which include dynamic two-stage training strategy, 3D-U-ResNet as the refiner network, global-aware fusion, and precision-recall loss function, are applied one by one based on it to convert the model to GARNet. Table~\ref{ablation_experiments} shows the results. Specifically, the setups of these experiments are as follows:
\begin{itemize}
\item[$\bullet$] \textbf{Setup 1}: Baseline (Pix2Vox++)
\end{itemize}
\begin{itemize}
\item[$\bullet$] \textbf{Setup 2}: Baseline + Dynamic two-stage training
\end{itemize}
\begin{itemize}
\item[$\bullet$] \textbf{Setup 3}: Baseline + Dynamic two-stage training+3D-U-Resnet
\end{itemize}
\begin{itemize}
\item[$\bullet$] \textbf{Setup 4}: Baseline + Dynamic two-stage training + 3D-U-Resnet + Globel-aware fusion
\end{itemize}
\begin{itemize}
\item[$\bullet$] \textbf{Setup 5}: Baseline + Dynamic two-stage training + 3D-U-Resnet + Globel-aware fusion + P-R loss (GARNet)
\end{itemize}
\begin{itemize}
\item[$\bullet$] \textbf{Setup 6}: Baseline + Dynamic two-stage training + 3D-U-Resnet + Globel-aware fusion + P-R loss + 8 views (GARNet+)
\end{itemize}

It verifies that these methods have a positive impact on the reconstruction algorithm. In addition, GARNet+ training with 8 views input performs well when facing a large number of views input. In this section, we will discuss the relevant verification works about pre-merger structure and training strategy in detail.

\subsubsection{Pre-Merger Block} 
We replace the fully connection layers with a 1D convolutional layer in the channel attention module of pre-merger block. Heavy parameters are not expected for a branch of weights calculation, because their lack of direct supervision will make training more difficult. In addition, channels of a vector are not strongly related to each other in the channel attention module. As a result, a simpler and lighter structure bring better performance, as shown in Table~\ref{bam+eca}.

The pre-merger block plays a critical role with only hundreds of parameters. It improves the performance of the model effectively by allowing data to flow across branches. To establish this connection is the most important responsibility of it, which increases countless potential network branches to promote the learning ability of the network for merging. For the feature representation, the decoder with a large number of parameters also assists it to generate a better global representation, so the network never lacks parameters but connections.
\begin{table}[!t]
	\centering
	\renewcommand\arraystretch{0.98}
	\scalebox{0.72}{
	\begin{tabular}{c|rrrrrrrrr}
		\textbf{Training Strategy} & 
		\multicolumn{1}{c}{\textbf{1 view}} & \multicolumn{1}{c}{\textbf{2 views}} & \multicolumn{1}{c}{\textbf{3 views}} & \multicolumn{1}{c}{\textbf{4 views}} & \multicolumn{1}{c}{\textbf{5 views}} \\ \hline
		\textbf{FASet\cite{yang2020robust}} & \textbf{0.6733} & 0.6994 & 0.7082 & 0.7126 & 0.7154 \\
		\textbf{Pix2Vox\cite{xie2019pix2vox}} & 0.6708 & 0.7019 & 0.7122 & 0.7173 & 0.7203 \\
		\textbf{Ours} & 0.6697 & \textbf{0.7023} & \textbf{0.7137} & \textbf{0.7195} & \textbf{0.7235} \\ \hline\hline
		\textbf{Training Strategy} & 
		 \multicolumn{1}{c}{\textbf{8 views}} & \multicolumn{1}{c}{\textbf{12 views}} & \multicolumn{1}{c}{\textbf{16 views}} & \multicolumn{1}{c}{\textbf{20 views}} & \\ \hline
		 \textbf{FASet} & 0.7195 & 0.7215 & 0.7228 & 0.7232 & \\
		\textbf{Pix2Vox} & 0.7250 & 0.7276 & 0.7290 & 0.7297 & \\
		\textbf{Ours} & \textbf{0.7289} & \textbf{0.7319} & \textbf{0.7336} & \textbf{0.7345} & \\ \hline
	\end{tabular}}
	\caption{Experiments on our architecture using only BCE loss function and setting the upper limit of input views to 3 during training to compare our proposed dynamic two-stage training strategy with the two existing strategies on ShapeNet using IoU.}
\label{training_strategy}
\end{table}

\subsubsection{Training Strategy}
\label{sec:training_strategy}

As aforementioned, the dynamic two-stage training strategy is more reasonable than the previous methods for the models using attention-based fusion. As shown in Table~\ref{training_strategy}, we employ these three strategies to train our network. As a result, our method has a distinct advantage in multi-view reconstruction tasks. For single-view reconstruction, our performance is slightly lower than theirs, but it is roughly the same. However, their models are overfitting for single-view reconstruction since training with single-view inputs sufficiently in the first stage, which is detrimental to the generalization ability of the model.

In addition, we explore the influence of the maximum number of input views during training for our strategy. Comparing GARNet and GARNet+, setting a higher upper limit with more memory consumption results in better performance on multi-view reconstruction, while losing the performance of single-view reconstruction due to the lower frequency of single-view input during training.

\begin{figure}[]
	\centering
	\scalebox{1.2}{
	\includegraphics[width=0.8\linewidth]{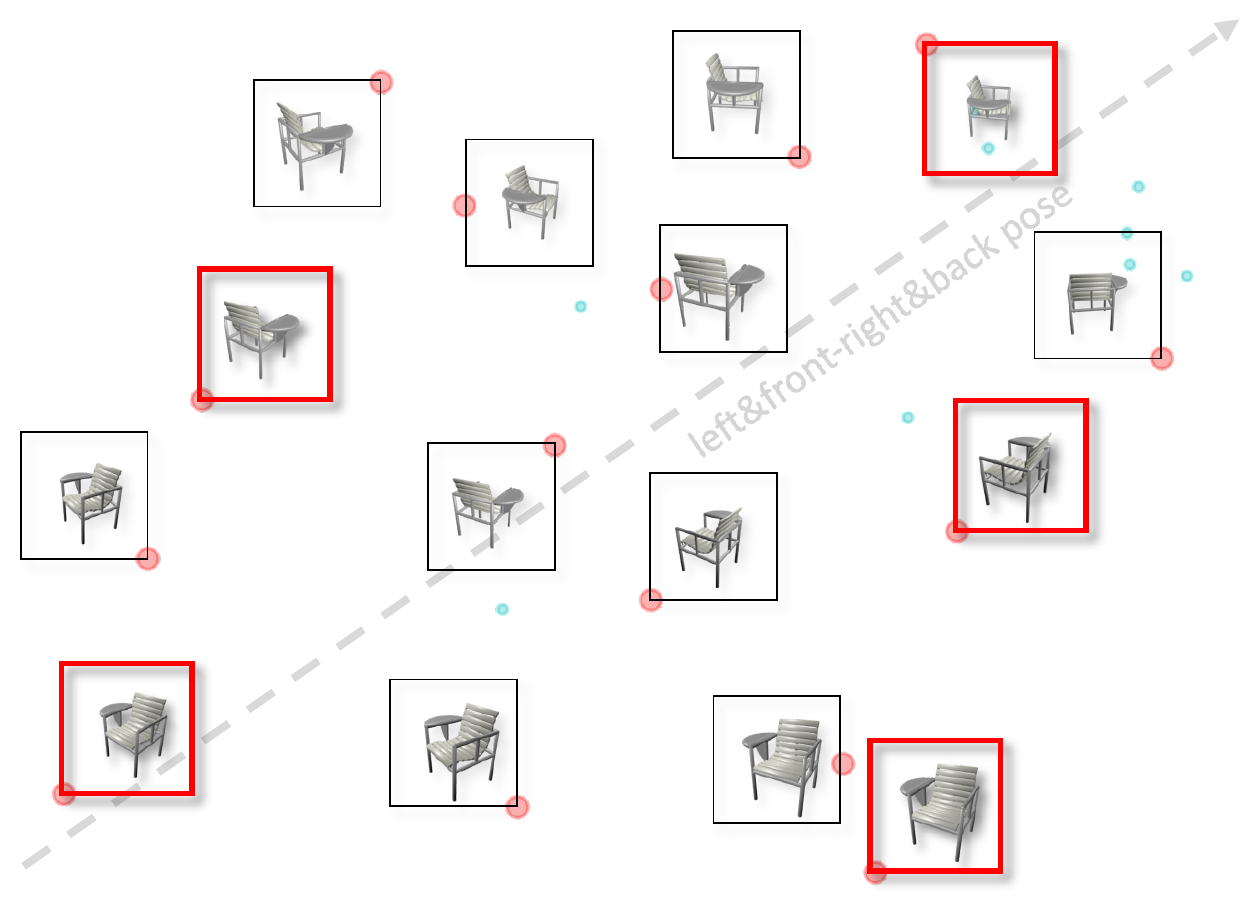}}
	\caption{The distribution of the points in the 2D manifold space mapped from the view images of an object by a trained IMB.}
\label{manifold_space}
\end{figure}

\begin{table}[]
	\centering
	\small
	\renewcommand\arraystretch{0.9}
	\scalebox{0.9}{
	\begin{tabular}{cc|ccc}
		\textbf{} & \multicolumn{1}{c|}{\textbf{View-Reduction}} & {\textbf{24} $\rightarrow$ \textbf{3}} & {\textbf{24} $\rightarrow$ \textbf{4}} & {\textbf{24} $\rightarrow$ \textbf{5}} \\ \hline
		& \textbf{Random} & 0.7163 & 0.7221 & 0.7254 \\
		& \textbf{FPS directly} & 0.7166 & 0.7223 & 0.7256 \\
		\multirow{-2}{*}{\textbf{GARNet}} & \textbf{PCA + FPS} & 0.7180 & 0.7243 & 0.7279 \\
		& \textbf{IMB + FPS (ours)} & \textbf{0.7225} & \textbf{0.7269} & \textbf{0.7295} \\ \hline
		& \textbf{Random} & 0.7120 & 0.7201 & 0.7246 \\
		& \textbf{FPS directly} & 0.7125 & 0.7205 & 0.7258 \\
		\multirow{-2}{*}{\textbf{GARNet+}} & \textbf{PCA + FPS} & 0.7138 & 0.7223 & 0.7281 \\
		& \textbf{IMB + FPS (ours)} & \textbf{0.7200} & \textbf{0.7263} & \textbf{0.7305} \\ \hline
	\end{tabular}}
	\caption{Comparison of the reconstruction performance based on the specified quantity of images reduced from 24 views by random sampling, FPS directly, PCA + FPS and our method, which is evaluated on test set of ShapeNet using IoU.}
\label{view-reduction}
\end{table}

\subsection{Cost-Performance Tradeoff}
\label{sec:tradeoff}

First of all, it is necessary to verify that our view-reduction method can retain the view combination including more diverse information. The different views of an object are mapped to a 2D manifold space by IMB and shown in Figure~\ref{manifold_space}. There is a certain correlation between the observation position of the chair and the location of the corresponding points in the space. It means that IMB extracts the features about viewpoints from the images. In addition, the 5 objects marked with red frames are selected by FPS and their viewpoints are obviously different from each other. Thus, view-reduction based on isometric mapping and diversity maximization is reliable. Note that, the displayed distribution is not ideally perfect, because the 2D vector cannot adequately capture the difference across viewpoints. It is merely for visualization clarity and we set a higher dimension (empirically set 5) for experiments.

In the experiments, we train IMB for GARNet and GARNet+ respectively. Table~\ref{view-reduction} illustrates that using our method can produce better results comparing to random sampling, using FPS directly or using FPS after PCA when reconstructing a volume using a specified number of images selected from 24 views. According to the statistics, using our view-reduction approach to retain 5 branches can achieve about 98\% precision of the results created based on all 24 views while saving about 70\% of computational cost.

\begin{figure}[]
    \centering
	\begin{minipage}{0.48\linewidth}
        \centerline{\includegraphics[width=1\linewidth]{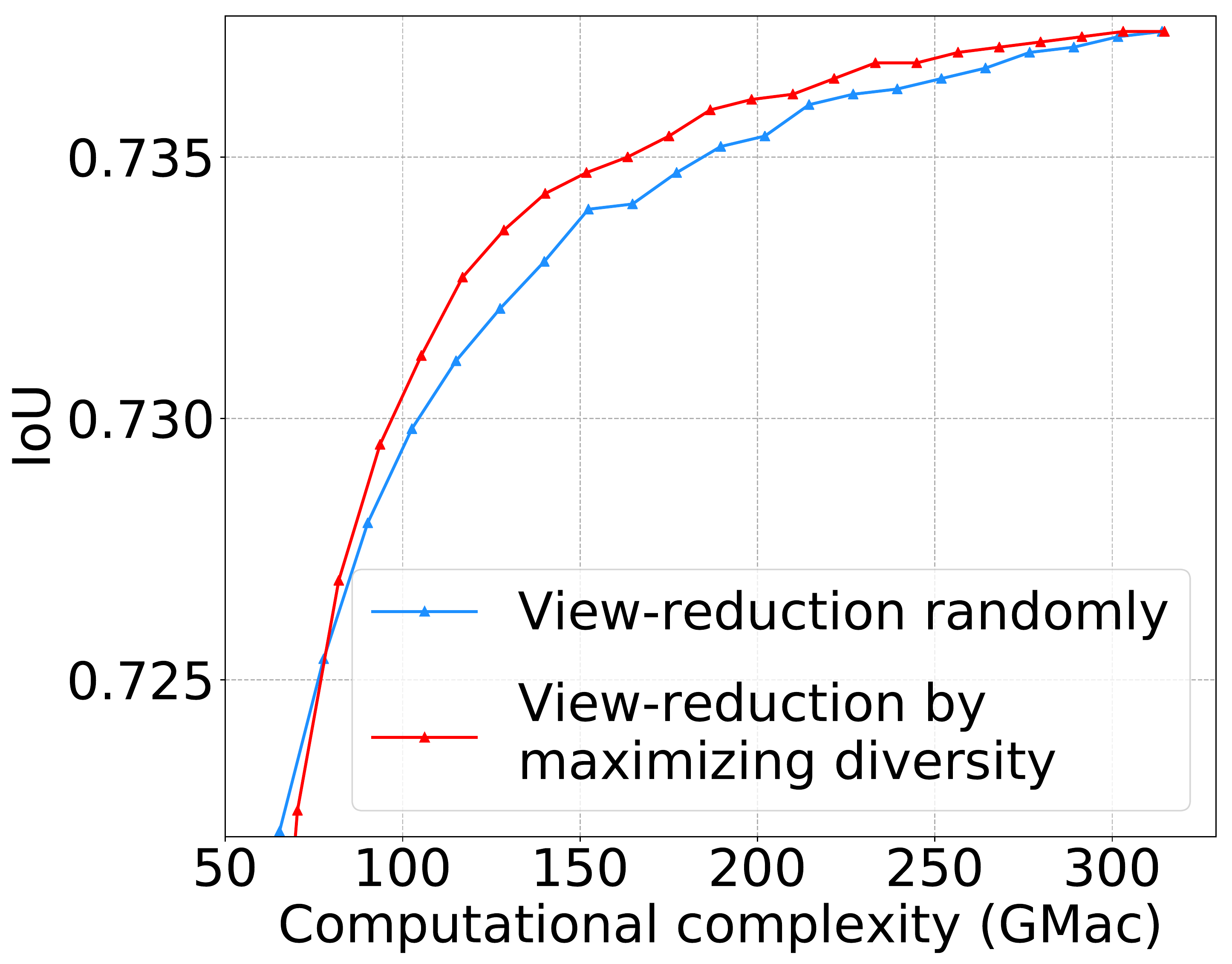}}
	\end{minipage}
	\begin{minipage}{0.48\linewidth}
        \centerline{ \includegraphics[width=1\linewidth]{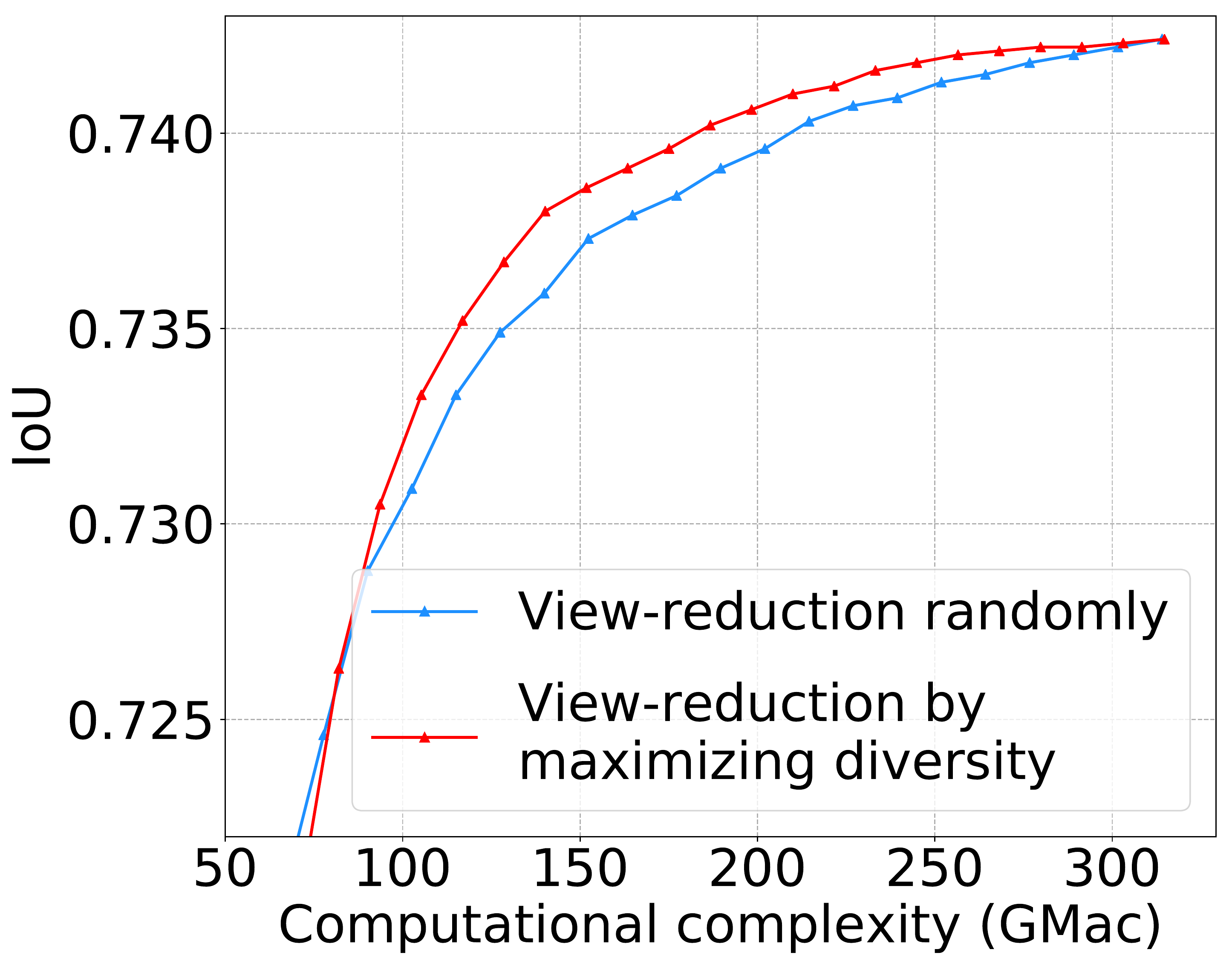}}
	\end{minipage}
	\caption{The cost-performance tradeoff that uses the maximizing diversity method and random sampling for view-reduction. The left one is the curve for GARNet and the right is for GARNet+.}
\label{tradeoff}
\end{figure}

For the cost-performance tradeoff, we have to analyze the relationship between specific computational cost and performance. When achieving the maximizing diversity selection, each view input must be processed by the network before the IMB. It is not required for view-reduction randomly. As a result, our method is not suitable for situations where the complexity is limited to a extremely low level. However, our view-reduction method still has applicable scenarios. Figure~\ref{tradeoff} presents the comparison of maximizing diversity and random sampling for view-reduction on GARNet and GARNet+. When the computational complexity greater than 80 GMac (multiply and accumulate) is allowed, our method takes advantage. In practical problems, such chart can help us determine a better course of action. Furthermore, the size of reconstructed volume is only $32^3$ with a low resolution because of the limitation of the dataset. If the model includes more decoder layers for a higher resolution result, view-reduction by maximizing diversity will obviously lead to a greater benefit.

\section{Conclusion}
In this paper, we propose a multi-view 3D reconstruction network using global-aware fusion, an advanced attention-based method, to favorably against the SOTA methods in performance. Furthermore, the cost-performance tradeoff is discussed for facing practical problems. In future work, we expect that these methods designed for the multi-branch network can be used to solve the multi-view reconstruction problem based on the other 3D representations and achieve high-resolution results.

{\small
\bibliographystyle{ieee_fullname}
\bibliography{3d_reconstruction}
}

\end{document}